\documentclass[10pt,twocolumn,letterpaper]{article}

\usepackage[pagenumbers]{iccv} 

\usepackage{xcolor}

\usepackage{graphicx}
\usepackage{caption}
\usepackage{amsmath}
\usepackage{multicol}
\usepackage{booktabs}
\usepackage{multirow}
\usepackage{rotating}
\usepackage{array}
\usepackage{amssymb}
\usepackage{colortbl}
\usepackage{siunitx}

\definecolor{iccvblue}{rgb}{0.21,0.49,0.74}
\usepackage[pagebackref,breaklinks,colorlinks,allcolors=iccvblue]{hyperref}

\def\paperID{} 
\def\confName{ICCV}
\def\confYear{2025}

\title{\textit{FaceXFormer}: A Unified Transformer for Facial Analysis}

\author{Kartik Narayan\thanks{Equal contribution} \quad Vibashan VS\footnotemark[1] \quad Rama Chellappa \quad Vishal M. Patel \\
{\tt\small \{knaraya4, vvishnu2, rchella4, vpatel36\}@jhu.edu}\\
\textcolor{magenta}{\small \url{https://kartik-3004.github.io/facexformer/}
}}

\begin{document}
\maketitle
\begin{abstract}
In this work, we introduce FaceXFormer, an end-to-end unified transformer model capable of performing ten facial analysis tasks within a single framework. These tasks include face parsing, landmark detection, head pose estimation, attribute prediction, age, gender, and race estimation, facial expression recognition, face recognition, and face visibility. Traditional face analysis approaches rely on task-specific architectures and pre-processing techniques, limiting scalability and integration. In contrast, FaceXFormer employs a transformer-based encoder-decoder architecture, where each task is represented as a learnable token, enabling seamless multi-task processing within a unified model. To enhance efficiency, we introduce FaceX, a lightweight decoder with a novel bi-directional cross-attention mechanism, which jointly processes face and task tokens to learn robust and generalized facial representations. We train FaceXFormer on ten diverse face perception datasets and evaluate it against both specialized and multi-task models across multiple benchmarks, demonstrating state-of-the-art or competitive performance. Additionally, we analyze the impact of various components of FaceXFormer on performance, assess real-world robustness in ``in-the-wild" settings, and conduct a computational performance evaluation. To the best of our knowledge, FaceXFormer is the first model capable of handling ten facial analysis tasks while maintaining real-time performance at $33.21$ FPS.
\end{abstract}
\section{Introduction}
\label{sec:intro}
\begin{figure}[t]
  \centering
   \includegraphics[width=0.47\textwidth]{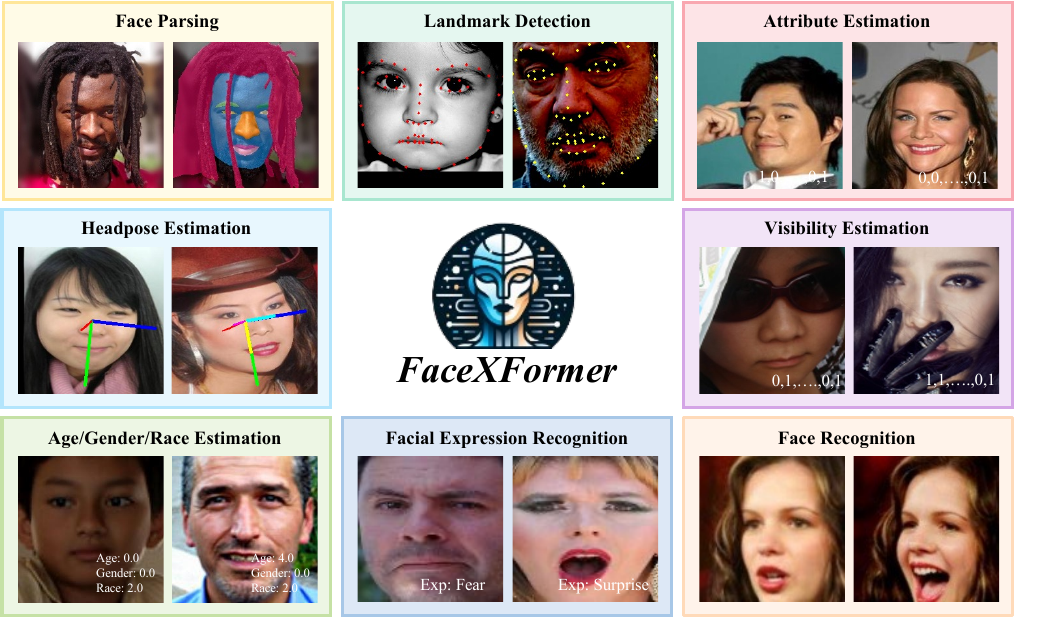}
   \caption{\textit{FaceXFormer} an end-to-end unified transformer model for 10 different facial analysis tasks such as face parsing, landmark detection, head pose estimation, attributes recognition, age, gender, and race estimation, facial expression recognition, face recognition, and face visibility prediction.}
   \label{fig:abstract}
    \vspace{-5pt}
\end{figure}

Face analysis is a crucial problem as it has broad range of application such as face verification and identification~\cite{sun2014deep, taigman2014deepface}, surveillance~\cite{ghalleb2020demographic}, face swapping~\cite{cui2023face}, face editing~\cite{zhu2020sean}, de-occlusion~\cite{yin2023segmentation}, 3D face reconstruction~\cite{wood20223d}, retail~\cite{abirami2020gender}, image generation~\cite{yan2016attribute2image} and face retrieval~\cite{zaeemzadeh2021face}. Facial analysis tasks (Figure~\ref{fig:abstract} include face parsing~\cite{jackson2016cnn, wei2017learning}, landmarks detection~\cite{lin2021structure, zhou2023star}, head pose estimation~\cite{zhou2020whenet, cobo2024representation}, facial attributes recognition~\cite{noroozi2016unsupervised, miyato2018virtual}, age/gender/race estimation~\cite{cao2020rank, li2021learning}, facial expression recognition~\cite{she2021dive}, face recognition~\cite{kim2022adaface}, and face visibility prediction~\cite{9523977, kumar2020luvli}. Therefore, developing a generalized and robust face model for all tasks is a crucial and longstanding problem in the face community.

\textbf{Why Unified Model ?} In recent years, significant advancements have been made in facial analysis, developing state-of-the-art methods and face libraries for various tasks ~\cite{zhou2020whenet,zhou2023star,li2021learning,cobo2024representation,yin2023segmentation,cui2023face}. Despite these methods achieving promising performance, they cannot be integrated into a single pipeline due to their specialized model designs and task-specific pre-processing techniques. Furthermore, deploying multiple specialized models simultaneously is computationally intensive and impractical for real-time applications, leading to increased system complexity and resource consumption. These challenges emphasis the need for a unified model that can concurrently handle multiple facial analysis tasks efficiently (see Table~\ref{tab:intro_table}). A single model capable of addressing multiple facial tasks is desirable because it: (1) learns a robust and generalized face representation capable of handling ``in-the-wild'' images; (2) intra-task modeling helps the models to learn task-invariant representation; and (3) simplifies deployment pipelines by reducing computational overhead and enabling faster inference.

\begin{table}[t]
    \centering
    \resizebox{1.0\columnwidth}{!}{
        \begin{tabular}{@{}lc@{\hspace{1em}}c@{\hspace{1em}}c@{\hspace{1em}}c@{\hspace{1em}}c@{\hspace{1em}}c@{\hspace{1em}}c@{\hspace{1em}}c@{\hspace{1em}}c@{\hspace{1em}}c@{}}
        \toprule
        Methods & FP & LD & HPE & Attr & Age & Gen & Race & Vis & Exp & FR\\ \midrule 
        \rowcolor{yellow!20}
        \multicolumn{11}{c}{\textbf{Single-Task Models}} \\
        DML-CSR~\cite{zheng2022decoupled} & \checkmark &  &  &  &  &  &  &  & &\\
        FP-LIIF~\cite{sarkar2023parameter} & \checkmark &  &  &  &  &  &  &  & &\\
        SegFace~\cite{narayan2024segface} & \checkmark &  &  &  &  &  &  &  & &\\
        Wing~\cite{Feng_2018_CVPR} &  & \checkmark &  &  &  &  &  &  & &\\
        HRNet~\cite{wang2020deep} &  & \checkmark &  &  &  &  &  &  & &\\
        WHENet~\cite{zhou2020whenet} &  &  & \checkmark &  &  &  &  &  & &\\
        TriNet~\cite{cao2021vector} &  &  & \checkmark &  &  &  &  &  & &\\
        img2pose~\cite{albiero2021img2pose} &  &  & \checkmark &  &  &  &  &  & &\\
        TokenHPE~\cite{zhang2023tokenhpe} &  &  & \checkmark &  &  &  &  &  & &\\ 
        SSPL~\cite{shu2021learning} &  &  &  & \checkmark &  &  &  &  & &\\
        VOLO-D1~\cite{kuprashevich2023mivolo} &  &  &  &  & \checkmark &  &  &  & &\\
        DLDL-v2~\cite{gao2020learning} &  &  &  &  & \checkmark &  &  &  & &\\
        3DDE~\cite{valle2019face} &  &  &  &  &  &  &  & \checkmark & &\\
        MNN~\cite{valle2020multi} &  &  &  &  &  &  &  & \checkmark & &\\
        KTN~\cite{li2021adaptively} &  &  &  &  &  &  &  &  & \checkmark &\\
        DMUE~\cite{she2021dive} &  &  &  &  &  &  &  &  & \checkmark &\\
        CosFace~\cite{wang2018cosface} &  &  &  &  &  &  &  &  &  & \checkmark\\
        ArcFace~\cite{deng2019arcface} &  &  &  &  &  &  &  &  & & \checkmark\\
        AdaFace~\cite{kim2022adaface} &  &  &  &  &  &  &  &  & & \checkmark\\
        \midrule 
        \rowcolor{green!20}
        \multicolumn{11}{c}{\textbf{Multi-Task Models}} \\
        SSP+SSG~\cite{kalayeh2017improving} & \checkmark &  &  & \checkmark &  &  &  &  & \\
        Hetero-FAE~\cite{han2017heterogeneous} &  &  &  & \checkmark & \checkmark & \checkmark & \checkmark &  & \checkmark\\
        FairFace~\cite{karkkainen2021fairface} &  &  &  &  & \checkmark & \checkmark & \checkmark &  & \\
        MiVOLO~\cite{kuprashevich2023mivolo} &  &  &  & & \checkmark & \checkmark & &  & \\
        MTL-CNN~\cite{zhuang2018multi} &  & \checkmark &  & \checkmark &  &  &  &  & \checkmark\\
        ProS~\cite{di2024pros} & \checkmark & \checkmark &  & \checkmark &  &  &  &  & \\
        FaRL~\cite{zheng2022general} & \checkmark & \checkmark &  & \checkmark & \checkmark & \checkmark &  &  & \\
        HyperFace~\cite{ranjan2017hyperface} &  & \checkmark & \checkmark &  &  & \checkmark &  & \checkmark & & \checkmark\\
        AllinOne~\cite{ranjan2017all} &  & \checkmark & \checkmark &  & \checkmark & \checkmark &  &  & \checkmark & \checkmark\\
        Swinface~\cite{qin2023swinface} &  & &  & \checkmark & \checkmark &  &   & & \checkmark & \checkmark\\
        QFace~\cite{sun2024task} &  &  & \checkmark  & \checkmark & \checkmark &  &  & & \checkmark & \\
        Faceptor~\cite{qin2025faceptor} & \checkmark & \checkmark &  & \checkmark & \checkmark & \checkmark &  & & \checkmark & \checkmark\\ \midrule
        \textbf{\textit{FaceXFormer}} & \checkmark & \checkmark & \checkmark & \checkmark & \checkmark & \checkmark & \checkmark & \checkmark & \checkmark & \checkmark\\ 
        \bottomrule
        \end{tabular}}
    \caption{Comparison with representative methods under different task settings. Our proposed \textit{FaceXFormer} can perform various facial analysis tasks in a single model. FP - Face Parsing, LD - Landmarks Detection, HPE - Head Pose Estimation, Attr - Attributes Recognition, Age - Age, Gen - Gender, Race - Race Estimation, Exp - Facial Expression Recognition, FR - Face Recognition, and Vis - Face Visibility Prediction  }
    \label{tab:intro_table}
    \vspace{-8pt}
\end{table}
\textbf{Proposed \textit{FaceXFormer} Architecture: }To this end, we introduce \textit{FaceXFormer}, an end-to-end unified model designed for ten different facial analysis tasks, as depicted in Figure~\ref{fig:abstract}. These tasks include face parsing, landmark detection, head pose estimation, attributes recognition, age/gender/race estimation, facial expression recognition, face recognition and face visibility prediction. \textit{FaceXFormer} enables task unification by leveraging transformers and learnable tokens as its core components. Specifically, we employ a transformer-based encoder-decoder structure, where the encoder extracts hierarchical face representations and fuses them using a MLP fusion module. The fused features are then processed in the decoder, where each facial analysis task is represented by a unique learnable token, allowing for the simultaneous and effective processing of multiple tasks. In particular, we propose a lightweight decoder, \textit{FaceX}, which processes both face and task tokens together using bi-directional cross-attention mechanism (Section~\ref{sec:decoder}), enabling the model to learn robust face representations that generalize across various tasks. The bi-directional cross-attention mechanism enables a 2-layer lightweight decoder, allowing the model to operate in real time. After modeling the intra-task and face-token relationships in the \textit{FaceX} decoder, the task tokens are fed into a unified head, which converts these task tokens into corresponding task predictions.

Our extensive experiments demonstrate that \textit{FaceXFormer} achieves state-of-the-art or competitive performance compared to specialized models and existing multi-task models across multiple benchmarks, while supporting more tasks than any previous multi-task model. Moreover, we show that our model effectively handles images ``in the wild'', demonstrating its robustness and generalizability across ten different tasks. This robustness is critical for real-world applications where uncontrolled conditions and diverse inputs are common. \textit{FaceXFormer} achieves state-of-the-art performance at $33.21$ FPS, representing a significant $69.44$\% speed boost over prior multi-task models, making it highly suitable for real-world applications.

In summary, our paper's contributions are as follows:
\begin{enumerate}
    \item We introduce \textit{FaceXFormer}, a unified transformer-based framework capable of simultaneously processing ten different facial analysis tasks, achieving real-time performance of $33.21$ FPS.
    \item We propose \textit{FaceX}, a lightweight decoder that employs the proposed bi-directional cross-attention mechanism, enabling joint processing of face and task tokens.
    \item We conduct extensive experiments and analyses to demonstrate that our approach achieves state-of-the-art performance with reduced inference time compared to specialized and multi-task models across multiple tasks.
\end{enumerate}

\section{Related Work}
\label{sec:related_work}
\noindent \textbf{Facial analysis tasks:} Facial analysis tasks involve face parsing~\cite{jackson2016cnn, chen2016attention, zheng2022decoupled, narayan2024segface}, landmarks detection~\cite{lin2021structure, zhou2023star, micaelli2023recurrence}, head pose estimation~\cite{zhou2020whenet, valle2020multi, zhang2023tokenhpe, cobo2024representation}, facial attributes recognition~\cite{noroozi2016unsupervised, miyato2018virtual, shu2021learning, zheng2022general}, age/gender/race estimation~\cite{cao2020rank, kuprashevich2023mivolo, levi2015age, li2021learning}, facial expression recognition~\cite{li2017reliable, li2021adaptively}, face recognition~\cite{wang2018cosface, deng2019arcface} and face visibility prediction~\cite{9523977, kumar2020luvli}. These tasks hold significance in various applications such as face swapping~\cite{cui2023face, narayan2023df}, face editing~\cite{zhu2020sean}, de-occlusion~\cite{yin2023segmentation}, 3D face reconstruction~\cite{wood20223d}, driver assistance~\cite{murphy2007head}, human-robot interaction~\cite{strazdas2021robo}, retail~\cite{abirami2020gender}, face verification and identification~\cite{sun2014deep, taigman2014deepface}, image generation~\cite{yan2016attribute2image}, image retrieval~\cite{zaeemzadeh2021face} and surveillance~\cite{ghalleb2020demographic, narayan2024petalface}. Specialized models excel in their respective tasks but cannot be easily integrated with other tasks due to the need for extensive task-specific pre-processing~\cite{lin2019face, zhou2023star}. Generally, these models under-perform when applied to tasks beyond their specialization as their design is specific to their designated tasks. Some works~\cite{zhang2014facial, ming2019dynamic, zhao2021deep, 7952703} perform multiple tasks simultaneously but utilize the additional tasks for guidance or auxiliary loss calculation to enhance the performance of the primary task. 

\noindent \textbf{Multi-task learning for face analysis:} HyperFace~\cite{ranjan2017hyperface} and AllinOne~\cite{ranjan2017all} are early convolution-based models that aim to perform multiple tasks. Recent multi-task frameworks, such as QFace~\cite{sun2024task} and Faceptor~\cite{qin2025faceptor}, are also inspired from DETR~\cite{carion2020end} and propose a unified model structure consisting of learnable tokens. However, these previous works differ from the proposed method in several key aspects, as summarized in Table~\ref{tab:related}. Specifically, QFace~\cite{sun2024task} employs a feature fusion module that uses stage embeddings to aggregate features from the encoder. Faceptor~\cite{qin2025faceptor} introduces a layer-attention mechanism to fuse features from different encoder layers and incorporates two separate decoders. Both methods, employ a 9-layer transformer decoder, also Faceptor additionally includes a Pixel Decoder. These architectural components increase computational overhead, resulting in slower inference times. In contrast, \textit{FaceXFormer} proposes a bi-directional cross-attention mechanism, which enables efficient task-specific feature extraction from face tokens resulting in a 2-layer lightweight decoder. This design choice is the primary reason for \textit{FaceXFormer}'s superior speed and performance. Notably, unlike previous methods, \textit{FaceXFormer} does not rely on face-specific pertaining backbone.
\begin{table}[t]
    \centering
    \resizebox{0.47\textwidth}{!}{
        \begin{tabular}{@{}lll>{\columncolor[HTML]{EFEFEF}}l}
            \toprule
            & \textbf{QFace}~\cite{sun2024task} & \textbf{Faceptor}~\cite{qin2025faceptor} & \textbf{\textit{FaceXFormer}} \\
            \midrule
            \textbf{Tasks} & 4 & 7 & 10 \\
            \textbf{Model Size} & - & 178.9 M & 109.29 M \\
            \textbf{Face Pretrained} & Yes & Yes & No \\
            \textbf{FPS (fp32)} & - & 14.30 & 33.21 \\
            \textbf{Decoder} & Query2Label~\cite{liu2107query2label} & Pixel Decoder \& & FaceX \\
            & & Transformer Decoder & \\
            \textbf{Decoder Layers} & 9 & 9 & 2 \\
            \bottomrule
        \end{tabular}%
    }
    \caption{Comparison of multi-task face analysis methods.}
    \label{tab:related}
    \vspace{-10pt}
\end{table}

\noindent  \textbf{Unified transformer models:} In recent years, the rise of transformers~\cite{vaswani2017attention, dosovitskiy2020image} have paved the way for the unification of multiple tasks within a single architecture. Unified transformer architectures are being explored across various computer vision problems, including segmentation~\cite{li2024omg, zou2024segment}, visual question answering (VQA)~\cite{wang2022omnivl, yuan2021florence}, tracking~\cite{wang2023tracking, zhu2023visual}, detection~\cite{wang2023detecting}. While these models may not achieve state-of-the-art (SOTA) performance and may under-perform compared to specialized models on some tasks, they demonstrate competitive performance across a variety of tasks. Such unification efforts have led to the development of foundational models like SAM~\cite{kirillov2023segment}, CLIP~\cite{radford2021learning}, LLaMA~\cite{touvron2023llama}, GPT-3~\cite{brown2020language}, DALL-E~\cite{ramesh2021zero}, etc. However, these models are computationally intensive and not suitable for facial analysis applications that require real-time performance. Motivated by this challenge, we propose \textit{FaceXFormer}: the first lightweight, transformer-based model capable of performing multiple facial analysis tasks. It delivers real-time performance at $33.21$ FPS and can be seamlessly integrated into existing systems providing additional annotations for the person of interest.

\section{FaceXFormer}
\label{sec:proposed_work}

\begin{figure*}[t]
  \centering
  \includegraphics[width=1.0\textwidth]{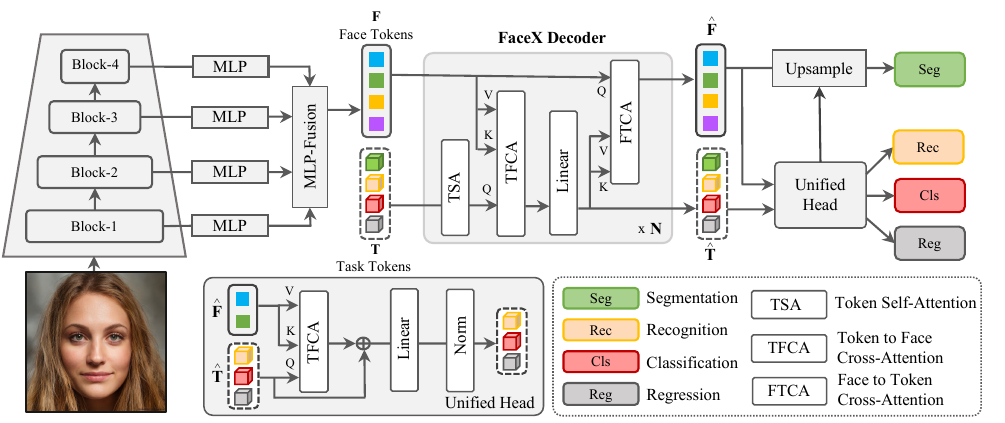}
  \vspace{-5pt}
   \caption{Overview of our proposed framework. The \textit{FaceXFormer} employs an encoder-decoder architecture, extracting multi-scale features from the input face image $\mathbf{I}$, and fusing them into a unified representation $\mathbf{F}$ via MLP-Fusion. Task tokens $\mathbf{T}$ are processed alongside face representation $\mathbf{F}$ in the FaceX Decoder $\mathbf{FXDec}$, resulting in refined task-specific tokens $\mathbf{\hat{T}}$. These refined tokens are then used for task-specific predictions by passing through the unified head. \textit{FaceXFormer} performs ten tasks, including face parsing, landmark detection, head pose estimation, attribute prediction, age, gender, and race estimation, facial expression recognition, face recognition, and face visibility prediction, achieving state-of-the-art performance at a real-time FPS of $33.21$. }
   \label{fig:archi}
   \vspace{-15pt}
\end{figure*}
In our framework, we follow a standard encoder-decoder structure as illustrated in Fig. \ref{fig:archi}. For an input face image $\mathbf{I} \in \mathbb{R}^{H \times W \times 3}$, we extract coarse to fine-grained multi-scale features $\mathbf{S}_i$, where $i$ belongs to the $i$-th encoder output. To learn a unified face representation  $\mathbf{F}$, these multi-scale features are then fused using a MLP-Fusion $\mathbf{M}$ module. Following fusion, we initialize a series of task-specific tokens $\mathbf{T} = \langle T_1, \dots, T_n \rangle$, with each $t_i$ representing a face task.
Afterward, we initialize task tokens $\mathbf{T} = \langle T_1, \dots, T_n \rangle$, where $T_i$ denotes each task. Face tokens $\mathbf{F}$ and task tokens $\mathbf{T}$ are then processed by a lightweight Decoder $\mathbf{FXDec}$ where task tokens are attended with face tokens to learn relevant task representation. 
\[
    \langle \mathbf{\hat{T}} \rangle = \mathbf{FXDec}\left(\langle\mathbf{F}, \mathbf{T}\rangle; \mathbf{S}_i\right)
    \label{EQ:FaceXformer_overall}
\]
Here, $\mathbf{\hat{T}}$ represents the output task tokens. These tokens are then fed into unified heads, where each task token is refined and passed to its respective task head for prediction.

\subsection{Multi-scale Encoder}
In the encoder, we employ a multi-scale encoding strategy to address the varying feature requirements intrinsic to each face analysis task. For instance, age estimation requires a global representation, while face parsing necessitates a fine-grained representation. Given an input image \(\mathbf{I}\), it is processed through a set of encoder layers. For each encoder layer, the output captures information at varying levels of abstraction and detail, generating multi-scale features \(\{\mathbf{S}_i\}_{i=1}^n\), where \(i\) ranges from 1 to 4. This results in a hierarchical structure of features, wherein each feature map \(\mathbf{S}_i\) transitions from a coarse to a fine-grained representation suitable for diverse facial analysis tasks.\\

\noindent \textbf{MLP-Fusion:} Assigning each feature-map \(\mathbf{S}_i\) to each face task is sub-optimal; rather, learning a unified face representation is more optimal and parameter-efficient. Following \cite{xie2021segformer}, we utilize a MLP-Fusion module \(\mathbf{M}\) to generate a fused face representation from the multi-scale features \(\{\mathbf{S}_i\}_{i=1}^n\). In this framework, each feature map \(\mathbf{S}_i\) is initially passed through a separate MLP layer, standardizing the channel dimensions across scales to facilitate fusion. The transformed features are then concatenated and passed through a fusion MLP layer to aggregate a fused representation \(\mathbf{F}\) as follows: 
\[
\begin{aligned}
    &\hat{\mathbf{S}}_i = \text{MLP}_{\text{proj}}(D_i, D_t)(\mathbf{S}_i), \forall i \in \{1, \ldots, n\},\\
    &\mathbf{F}_{\text{cat}} = \text{Concat}(\hat{\mathbf{S}}_1, \hat{\mathbf{S}}_2, \ldots, \hat{\mathbf{S}}_n),\\
    &\mathbf{F} = \text{MLP}_{\text{fusion}}(nD_t, D_t)(\mathbf{F}_{\text{cat}}),
\end{aligned}
\]
where \(D_i\) and \(D_t\) are the multi-scale feature channel dimensions of \(\mathbf{S}_i\) and the target channel dimension, respectively. The MLP-fusion design ensures minimal computational overhead (983k parameters) while maintaining the ability to perform efficient feature fusion, which is crucial for real-time application based face analysis tasks.

\subsection{FaceX Decoder}
\label{sec:decoder}
Detection transformer (DETR) \cite{carion2020end} employs object tokens to learn bounding box predictions for each object. Inspired by this approach, we introduce Task Tokens, whereby each task token is designed to learn specific facial tasks leveraging the fused face representation. However, existing decoders such as DETR \cite{carion2020end} and Deformable-DETR \cite{zhu2020deformable} are computationally intensive, impacting runtime significantly. To address this, we propose \textit{FaceX} ($\mathbf{FXDec}$) a lightweight decoder  designed to efficiently model the task tokens with face tokens. Specifically, each task token learns a task-related representation by interacting with other task tokens $\mathbf{T}$ and face tokens $\mathbf{F}$, enhancing the overall representation. The Lightweight Decoder comprises of three main components: 1) Task Self-Attention, 2) Task-to-Face Cross-Attention, and 3) Face-to-Task Cross-Attention as illustrated in Figure~\ref{fig:archi}.\\

\noindent \textbf{Task Self-Attention (TSA):}
The Task Self-Attention module is designed to refine the task-specific representations within the set of task tokens  $\mathbf{T} = \langle T_1, \dots, T_n \rangle$. Each task token $T_i$ is an embedded representation that corresponds to a specific facial task. In TSA, each $T_i$ is updated by attending to all other task tokens to capture task-specific interactions. Formally, the updated task token $T'_i$ is computed as:
\[
\mathbf{T}'_i = \text{SelfAttn}(\mathbf{Q}=T'_i, \mathbf{K}=\mathbf{T}, \mathbf{V}=\mathbf{T}),
\]
where $\text{Attention}$ denotes the multi-headed self-attention mechanism, and $\mathbf{Q}$, $\mathbf{K}$, and $\mathbf{V}$ represent the queries, keys, and values, respectively. Therefore, TSA essentially helps the model to learn task-invariant representation.\\

\noindent \textbf{Task-to-Face Cross-Attention (TFCA):}
The Task-to-Face Cross-Attention module allows each task token to interact with the fused face representation $\mathbf{F}$. This enables each task token to gather information relevant to its specific facial task from the fused face features. In this module, the fused face representation $\mathbf{F}$ acts as both key and value, while the task tokens serve as queries. The updated task token $\hat{T}_i$ is then computed as follows:
\[
\hat{T}_i = \text{CrossAttn}(\mathbf{Q}={T}'_i, \mathbf{K}=\mathbf{F}, \mathbf{V}=\mathbf{F}),
\]
where $\mathbf{\hat{T}} = \langle \hat{T}_1, \dots, \hat{T}_n \rangle$ is the output task token. Thus, TFCA enables direct interaction between the task-specific tokens and the compact facial features, facilitating task-focused feature extraction.\\

\noindent \textbf{Face-to-Task Cross-Attention (FTCA):}
Conversely, the Face-to-Task Cross-Attention module is designed to refine the fused face representation $\mathbf{F}$ based on the information from the updated task tokens. This process aids in enhancing the face representation with task-specific details, thereby improving the extraction of overall fused representation. In FTCA, the set of updated task tokens $\mathbf{T}' = \{\mathbf{T}''_1, \mathbf{T}''_2, \ldots, \mathbf{T}''_m\}$ acts as both keys and values, while the fused face features $\mathbf{F}$ serve as queries. The refined face representation $\mathbf{\hat{F}}$ is computed as:
\[
\mathbf{\hat{F}} = \text{CrossAttn}(\mathbf{Q}=\mathbf{F}, \mathbf{K}=\mathbf{T}', \mathbf{V}=\mathbf{T}').
\]
Through this inverse attention mechanism, the face representation is augmented with critical task-specific details, enabling a robust approach towards facial task unification.

\subsection{Unified-Head}
In Unified-Head, the task tokens are processed to obtain corresponding task predictions. As shown in Figure~\ref{fig:archi}, the output face tokens $\mathbf{\hat{F}}$ and task tokens $\mathbf{\hat{T}}$ are processed through a Task-to-Face Cross-Attention mechanism to obtain final refined features. Then, the output tokens are fed into their corresponding task heads. The task head for landmark detection is a hourglass network, for head pose estimation is a regression MLP, and for face recognition is PartialFC~\cite{an2021partial}, while the tasks of age, gender and race estimation, facial expression recognition, face visibility prediction, and attributes prediction utilize classification MLPs. For face parsing, we leverage the output $\mathbf{\hat{F}}$ and process it through an upsampling layer, then perform a cross-product with the face parsing token to obtain a segmentation map. The number of tokens for segmentation corresponds to the total number of classes. For landmark prediction, it corresponds to the number of landmarks (i.e., $68$). For head pose estimation, the number of tokens is $9$, representing the $3\times3$ rotation matrix. For other tasks, one token is used for each.

\subsection{Multi-Task Training}
We aim to train \textit{FaceXFormer} for multiple facial analysis tasks simultaneously, however each task requires distinct and sometimes conflicting pre-processing steps. For instance, landmark detection typically requires keypoint alignment of faces, which contradicts the needs for head pose estimation, as it may eliminate the natural variability of headposes. Due to these reasons, integrating all tasks into a single model poses significant challenges. To address this, \textit{FaceXFormer} incorporates task-specific tokens designed to extract task-specific features from the fused representation. These task tokens compel the backbone to learn a unified representation capable of supporting a broad spectrum of facial analysis tasks. We employ different loss functions for each task and combine them in a joint objective for training. The final loss function is given as:

\resizebox{0.45\textwidth}{!}{$
\begin{aligned}
L = \lambda_{seg}L_{seg} &+ \lambda_{lnd}L_{lnd} + \lambda_{hpe}L_{hpe} + \lambda_{attr}L_{attr} + \lambda_{a}L_{a} \\
& + \lambda_{g/r}L_{g/r} + \lambda_{exp}L_{exp} + \lambda_{fr}L_{fr} + \lambda_{vis}L_{vis}
\end{aligned}
$}

\noindent where $L_{seg}$ is the mean of dice loss~\cite{sudre2017generalised} and Cross-Entropy (CE) loss for face parsing, $L_{lnd}$ is STAR loss~\cite{zhou2023star} for landmarks prediction, $L_{hpe}$ is geodesic loss~\cite{zhang2023tokenhpe} for head pose estimation, $L_{g/r}$ is CE loss for gender/race estimation, $L_{a}$ is mean of L1 loss and CE loss for age estimation, $L_{exp}$ is CE loss for facial expression recognition, $L_{fr}$ is ArcFace~\cite{deng2019arcface} loss for face recognition, and $L_{attr}$ and $L_{vis}$ are Binary Cross-Entropy with logits loss for attributes prediction and face visibility prediction respectively. 
\begin{table*}[t]
    \centering
    \resizebox{1.0\textwidth}{!}{\begin{tabular}{@{}l c|c c c c c c c c c c |c@{}}
    \toprule
    \textbf{Method} & \textbf{Input Res.} &\textbf{Skin} & \textbf{Hair} & \textbf{Nose} & \textbf{L-Eye} & \textbf{R-Eye} & \textbf{L-Brow} & \textbf{R-Brow} & \textbf{L-Lip} & \textbf{I-Mouth} & \textbf{U-Lip} & \textbf{Mean F1} \\
    \midrule
    Wei et al.~\cite{wei2019accurate} & 512 & 96.4 & 91.1 & 91.9 & 87.1 & 85.0 & 80.8 & 82.5 & 91.0 & 90.6 & 87.9 & 88.43 \\
    EHANet~\cite{luo2020ehanet} & 512 & 96.0 & 93.9 & 93.7 & 86.2 & 86.5 & 83.2 & 83.1 & 90.3 & \textcolor{blue}{\underline{93.8}} & 88.6 & 89.53 \\
    EAGRNet~\cite{te2020edge} & 473 & 96.2 & 94.9 & \textcolor{red}{\textbf{94.0}} & 88.6 & 89.0 & 85.7 & 85.2 & \textcolor{red}{\textbf{91.2}} & \textcolor{red}{\textbf{95.0}} & 88.9 & 90.87 \\
    AGRNet~\cite{te2021agrnet} & 473 & \textcolor{blue}{\underline{96.5}} & 87.6 & 93.9 & 88.7 & 89.1 & 85.5 & 85.6 & 91.1 & 92.0 & 89.1 & 89.91 \\
    FaRL$_{\text{scratch}}$~\cite{zheng2022general} & 512 & 96.2 & 94.9 & 93.8 & 89.0 & 89.0 & 85.3 & 85.4 & 90.0 & 91.7 & 88.1 & 90.34\\
    DML-CSR~\cite{zheng2022decoupled} & 473 & 95.7 & 94.5 & \textcolor{blue}{\underline{93.9}} & \textcolor{blue}{\underline{89.4}} & \textcolor{blue}{\underline{89.6}} & 85.5 & 85.7 & \textcolor{blue}{\underline{91.0}} & 91.8 & \textcolor{blue}{\underline{89.1}} & 90.62\\
    FP-LIIF~\cite{sarkar2023parameter} & 256 & 96.4 & 95.1 & 93.7 & 88.5 & 88.5 & 84.5 & 84.3 & 90.3 & 92.1 & 87.5 & 90.09\\
    \midrule
    SwinFace~\cite{qin2023swinface} & $\times$ & $\times$ & $\times$ & $\times$ & $\times$ & $\times$ & $\times$ & $\times$ & $\times$ & $\times$ & $\times$ & $\times$\\
    QFace~\cite{sun2024task} & $\times$ & $\times$ & $\times$ & $\times$ & $\times$ & $\times$ & $\times$ & $\times$ & $\times$ & $\times$ & $\times$ & $\times$\\
    Faceptor~\cite{qin2025faceptor} & 512 & \textcolor{red}{\textbf{96.6}} & \textcolor{red}{\textbf{96.2}} & 93.9 & 89.4 & 89.1 & \textcolor{red}{\textbf{86.2}} & \textcolor{red}{\textbf{86.3}} & 90.6 & 91.6 & 89.0 & \textcolor{blue}{\underline{90.89}}\\ 
    \textbf{\textit{FaceXFormer}} & 224 & 96.4 & \textcolor{blue}{\underline{95.7}} & 93.8 & \textcolor{red}{\textbf{90.1}} & \textcolor{red}{\textbf{90.3}} & \textcolor{blue}{\underline{86.0}} & \textcolor{blue}{\underline{85.9}} & 90.6 & 92.1 & \textcolor{red}{\textbf{89.2}} & \textcolor{red}{\textbf{92.01}}\\ 
    \bottomrule
    \end{tabular}}
    \vspace{-5pt}
    \caption{Performance comparison for face parsing on the CelebAMask-HQ dataset~\cite{CelebAMask-HQ}. The symbol $\times$ indicates that the model does not perform the corresponding task. \textcolor{red}{\textbf{Red}} = First Best, \textcolor{blue}{\underline{Blue}} = Second Best. $\times$ indicates a model that doesn't perform the task.}
    \label{tab:seg}
    \vspace{-5pt}
\end{table*}
\vspace{-2pt}
\section{Experiments and Results}
\label{sec:experiments}

\subsection{Datasets and Metrics}
\label{subsec:datasets_metrics}
We perform co-training, where the model is simultaneously trained for multiple tasks using a total of 10 datasets with task-specific annotations. We conduct a comprehensive evaluation, comparing our approach with both task-specific and multi-task models. We present our results on the test sets according to the standard protocol for each task using the following datasets:\\
\noindent \textbf{Train:} \textit{Face Farsing}: CelebAMaskHQ~\cite{CelebAMask-HQ}; \textit{Landmarks Detection}: 300W~\cite{sagonas2013300}; \textit{Head Pose Estimation}: 300W-LP~\cite{zhu2016face}; \textit{Attributes Prediction}: CelebA~\cite{liu2015faceattributes}; \textit{Facial Expression Recognition}: RAF-DB~\cite{li2017reliable}, AffectNet~\cite{mollahosseini2017affectnet}; \textit{Age/Gender/Race estimation}: UTKFace~\cite{zhang2017age}, FairFace~\cite{karkkainenfairface}; \textit{Face Recognition}: MS1MV3~\cite{guo2016ms}; \textit{Visibility Prediction}: COFW~\cite{burgos2013robust}. \\
\noindent \textbf{Test:} \textit{Face Parsing}: CelebAMaskHQ~\cite{CelebAMask-HQ}; \textit{Landmarks Detection}: 300W~\cite{zhu2016face}, 300VW~\cite{shen2015first}; \textit{Head Pose Estimation}: BIWI~\cite{fanelli2013random}; \textit{Attributes Prediction}: CelebA~\cite{liu2015faceattributes}, LFWA~\cite{wolf2010effective}; \textit{Facial Expression Recognition}: RAF-DB~\cite{li2017reliable}; \textit{Age/Gender/Race Estimation}: UTKFace~\cite{zhang2017age}, FairFace~\cite{karkkainenfairface}; \textit{Face Recognition}: LFW~\cite{huang2008labeled}, CFP-FP~\cite{cfp-paper}, AgeDB~\cite{moschoglou2017agedb}, CALFW~\cite{zheng2017cross}, CPLFW~\cite{zheng2018cross} ; \textit{Visibility Prediction}: COFW~\cite{burgos2013robust}. 

\noindent The evaluation metrics used are the F1-score for face parsing, Normalized Mean Error (NME) for landmark prediction, Mean Absolute Error (MAE) for head pose estimation and age estimation, accuracy for facial expression recognition, attributes prediction, gender estimation, race estimation, 1:1 verification accuracy for face recognition, and recall at $80$\% precision for face visibility prediction.

\begin{table}[ht]
\centering
\resizebox{0.47\textwidth}{!}{
\begin{tabular}{@{}l c | l c | l c@{}}
\toprule
\multirow{2}{*}{\textbf{Methods}} & \textbf{Expression} & \multirow{2}{*}{\textbf{Methods}} & \textbf{Visibility} & \multirow{2}{*}{\textbf{Methods}} & \textbf{Age (MAE)} \\
& \textbf{(RAF-DB)} & & \textbf{(COFW)} & & \textbf{UTKFace} \\
\midrule
DLP-CNN~\cite{li2017reliable} & 80.89 & RCPR~\cite{burgos2013robust} & 40 & OR-CNN~\cite{niu2016ordinal} & 5.74  \\
gACNN~\cite{li2018occlusion} & 85.07 & Wu et al.~\cite{wu2017simultaneous} & 44.43 & Axel Berg et al.~\cite{berg2021deep} & 4.55 \\
IPA2LT~\cite{zeng2018facial} & 86.77 & Wu et al.~\cite{wu2015robust} & 49.11 & CORAL~\cite{cao2020rank} & 5.47 \\
RAN~\cite{wang2020region} & 86.90 & ECT~\cite{zhang2018combining} & 63.4 & Gustafsson et al.~\cite{gustafsson2020energy} & 4.65 \\
CovPool~\cite{acharya2018covariance} & 87.00 & 3DDE~\cite{valle2019face} & 63.89 & R50-SORD~\cite{paplham2024call} & 4.36 \\
SCN~\cite{wang2020suppressing} & 87.03 & MNN~\cite{valle2020multi} & \textcolor{blue}{\underline{72.12}} & VOLO-D1~\cite{kuprashevich2023mivolo} & 4.23 \\
DACL~\cite{farzaneh2021facial} & 87.78 & & & DLDL-v2~\cite{gao2020learning} & 4.42 \\
KTN~\cite{li2021adaptively} & 88.07 & & & MWR~\cite{shin2022moving} & 4.37 \\
DMUE~\cite{she2021dive} & \textcolor{blue}{\underline{88.76}} & & & \\
\midrule
SwinFace~\cite{qin2023swinface} & 86.54 & SwinFace~\cite{qin2023swinface} & $\times$ & SwinFace~\cite{qin2023swinface} & $\times$ \\
QFace~\cite{sun2024task} & \textcolor{red}{\textbf{92.86}} & QFace~\cite{sun2024task} & $\times$ & QFace~\cite{sun2024task} & $\times$  \\
Faceptor~\cite{qin2025faceptor} & 87.58 & Faceptor~\cite{qin2025faceptor} & $\times$ & Faceptor~\cite{qin2025faceptor} & \textcolor{red}{\textbf{4.10}}  \\
\textbf{\textit{FaceXFormer}} & \textbf{88.24} & \textbf{\textit{FaceXFormer}} & \textcolor{red}{\textbf{72.56}} & \textbf{\textit{FaceXFormer}} & \textcolor{blue}{\underline{4.17}} \\
\bottomrule
\end{tabular}
}
\vspace{-5pt}
\caption{Performance comparison on facial expression recognition, face visibility prediction and age estimation.}
\label{tab:exp}
\vspace{-10pt}
\end{table}

\subsection{Implementation Details}
\label{subsec:implementation}

We train our models using a distributed PyTorch setup on eight A6000 GPUs, each equipped with $48$GB of memory. The models' backbones are initialized with ImageNet pre-trained weights and processes input images at a resolution of $224 \times 224$. We employ the AdamW optimizer with a weight decay of $1e^{-5}$. All models are trained for $12$ epochs with a batch size of $48$ on each GPU, and an initial learning rate of $1e^{-4}$, which decays by a factor of $10$ at the $6^{th}$ and $10^{th}$ epochs. We train the model for three additional epochs for some tasks. For data augmentation, we randomly apply Gaussian blur, grayscale conversion, gamma correction, occlusion, horizontal flipping, and affine transformations, such as rotation, translation and scaling. The number of \textit{FaceX} decoder $N$ is set to two. To ensure stable training across tasks when using multiple datasets of varying sample sizes, we equalize the representation of each task's samples in every batch through upsampling. Additional details on our implementation are provided in the Appendix~\ref{sec:supp_implementation}.

\subsection{Main results}
\label{subsec:results}
\begin{table*}[ht]
    \centering
    \centering
\resizebox{0.97\textwidth}{!}{\begin{tabular}{@{}l cccc | l ccc | l c@{}}
        \toprule
        \multirow{2}{*}{\textbf{Methods}} & \multicolumn{4}{c|}{\textbf{Headpose (BIWI)}} & \multirow{2}{*}{\textbf{Methods}} & \multicolumn{3}{c|}{\textbf{Landmarks (300W)}} & \multirow{2}{*}{\textbf{Methods}} & \multicolumn{1}{c}{\textbf{CelebA}} \\
        & \textbf{Yaw} & \textbf{Pitch} & \textbf{Roll} & \textbf{MAE} & & \textbf{Full} & \textbf{Com} & \textbf{Chal} & & \textbf{Acc.} \\
        \midrule
        HopeNet~\cite{ruiz2018fine} & 4.81 & 6.61 & 3.27 & 4.89 & LAB~\cite{wu2018look} & 3.49 & 2.98 & 5.19 & PANDA-1~\cite{zhang2014panda} & 85.43 \\
        QuatNet~\cite{hsu2018quatnet} & 5.49 & 4.01 & 2.94 & 4.15 & Wing~\cite{Feng_2018_CVPR} & 4.04 & 3.27 & 7.18 & LNets+ANet~\cite{liu2015deep} & 87.33 \\
        FSA-Net~\cite{yang2019fsa} & 4.27 & 5.49 & 2.93 & 4.14 & DeCaFa~\cite{dapogny2019decafa} & 3.39 & 2.93 & 5.26 & SSP+SSG~\cite{kalayeh2017improving} & 88.24 \\
        EVA-GCN~\cite{xin2021eva} & 6.01 & 4.78 & 2.98 & 3.98 & HRNet~\cite{wang2020deep} & 3.32 & 2.87 & 5.15 & MOON~\cite{rudd2016moon} & 90.94 \\
        TriNet~\cite{cao2021vector} & 4.11 & 4.75 & 3.04 & 3.97 & PicassoNet~\cite{9764821} & 3.58 & 3.03 & 5.81 & NSA~\cite{mahbub2018segment} & 90.61 \\
        img2pose~\cite{albiero2021img2pose} & 4.56 & \textcolor{red}{\textbf{3.54}} & 3.24 & 3.78 & AVS+SAN~\cite{dong2018style} & 3.86 & 3.21 & 6.46 & MCNN-AUX~\cite{hand2017attributes} & 91.29 \\
        MNN~\cite{valle2020multi} & 3.98 & 4.61 & \textcolor{red}{\textbf{2.39}} & 3.66 & LUVLi~\cite{kumar2020luvli} & 3.23 & 2.76 & 5.16 & MCFA~\cite{zhuang2018multi} & 91.23 \\
        MFDNet~\cite{liu2021mfdnet} & \textcolor{red}{\textbf{3.40}} & 4.68 & 2.77 & \textcolor{blue}{\underline{3.62}} & HIH~\cite{lan2021hih} & \textcolor{blue}{\underline{3.09}} & \textcolor{red}{\textbf{2.65}} & 4.89 & DMM-CNN~\cite{mao2020deep} & 91.70 \\
        TokenHPE~\cite{zhang2023tokenhpe} & 3.95 & 4.51 & 2.71 & 3.72 & PIPNet~\cite{jin2021pixel} & 3.19 & 2.78 & 4.89 & SSPL~\cite{shu2021learning} & \textcolor{blue}{\underline{91.77}} \\
        WHENet~\cite{zhou2020whenet} & 3.99 & 4.39 & 3.06 & 3.81 & SLPT~\cite{xia2022sparse} & 3.17 & 2.75 & 4.90 & FaRL~\cite{zheng2022general} & 91.39 \\
        \midrule
        SwinFace~\cite{qin2023swinface} & $\times$ & $\times$ & $\times$ & $\times$ & SwinFace~\cite{qin2023swinface} & $\times$ & $\times$ & $\times$ & SwinFace~\cite{qin2023swinface} & 91.38 \\
        QFace~\cite{sun2024task} & -- & -- & -- & -- & QFace~\cite{sun2024task} & $\times$ & $\times$ & $\times$ & QFace~\cite{sun2024task} & 91.56 \\
        Faceptor~\cite{qin2025faceptor} & $\times$ & $\times$ & $\times$ & $\times$  & Faceptor~\cite{qin2025faceptor} & 3.16 & 2.75 & \textcolor{blue}{\underline{4.84}} & Faceptor~\cite{qin2025faceptor} & 91.39 \\
        \textit{\textbf{FaceXFormer}} & \textcolor{blue}{\underline{3.91}} & \textcolor{blue}{\underline{3.97}} & \textcolor{blue}{\underline{2.67}} & \textcolor{red}{\textbf{3.52}} & \textit{\textbf{FaceXFormer}} & \textcolor{red}{\textbf{3.05}} & \textcolor{blue}{\underline{2.66}} & \textcolor{red}{\textbf{4.67}} & \textit{\textbf{FaceXFormer}} & \textcolor{red}{\textbf{91.83}} \\
        \bottomrule
    \end{tabular}}
    \vspace{-5pt}
    \caption{Performance comparison on headpose, landmark detection, and attribute recognition. The symbol $\times$ indicates that the model does not perform the corresponding task, while -- denotes that results for this dataset are not provided. \textcolor{red}{\textbf{Red}} = First Best, \textcolor{blue}{\underline{Blue}} = Second Best.}
    \vspace{-10pt}
    \label{tab:hpe}
\end{table*}

In Table~\ref{tab:seg}, Table~\ref{tab:exp}, Table~\ref{tab:fr}, Table~\ref{tab:hpe}, we present a comparative analysis of \textit{FaceXFormer} against recent methods across a variety of tasks. A key highlight of our work is its unique capability to deliver promising results across multiple tasks at real-time inference speed using a single unified model. Specifically, \textit{FaceXFormer} achieves state-of-the-art performance in face parsing, with a mean F1 score of $92.01$ on CelebAMaskHQ at a resolution of $224 \times 224$, which is half the input size required by other state-of-the-art methods. Furthermore, it demonstrates superior performance in head pose estimation and landmark detection, achieving a mean MAE of $3.52$ and a mean NME of $4.67$, respectively. Additionally, \textit{FaceXFormer} provides a significant performance boost in attributes prediction and visibility prediction, achieving an accuracy of $91.83$\% on the CelebA dataset and $72.56$\% on COFW. It also performs competitively in age estimation, achieving the second-best score of $4.17$, and achieves an accuracy of $88.24$\% in facial expression recognition. In face recognition, \textit{FaceXFormer} outperforms Faceptor, achieving a mean accuracy of $95.94$\% compared to $95.28$\%. However, we observe that multi-task models generally underperform compared to specialized ones in this task. This can be attributed to conflicting training objectives, which force the model to learn identity-invariant features rather than identity-specific representations crucial for accurate recognition. The results on gender estimation across different race categories is shown in Table~\ref{tab:ethics}. We present additional cross-dataset results in Appendix~\ref{sec:supp_cross}.

\begin{table}[ht]
    \centering
    \resizebox{0.47\textwidth}{!}{%
        \begin{tabular}{lcccccc}
            \toprule
             \textbf{Method} & \textbf{LFW} & \textbf{CFP-FP} & \textbf{AgeDB} & \textbf{CALFW} & \textbf{CPLFW}& \textbf{Mean} \\ \midrule
             CosFace~\cite{wang2018cosface} & 99.81 & 98.12 & 98.11 & 95.76 & 92.28 & 96.81 \\
             ArcFace~\cite{deng2019arcface} & 99.83 & 98.27 & \textcolor{blue}{\underline{98.28}} & 95.45 & 92.08 & 96.78 \\ 
             VPL-ArcFace~\cite{deng2021variational}           & 99.83        & \textcolor{blue}{\underline{99.11}}          & \textcolor{red}{\textbf{98.60}}             & \textcolor{red}{\textbf{96.12}}          & \textcolor{blue}{\underline{93.45}}          & \textcolor{red}{\textbf{97.42}}                     \\
             AdaFace~\cite{kim2022adaface}           & \textcolor{blue}{\underline{99.83}}        & \textcolor{red}{\textbf{99.11}}          & 98.17             & 96.02          & \textcolor{red}{\textbf{93.93}}          & \textcolor{blue}{\underline{97.41}}                     \\ 
             \midrule
             SwinFace~\cite{qin2023swinface}       & \textcolor{red}{\textbf{99.87}}        & 98.60          & 98.15             & \textcolor{blue}{\underline{96.10}}          & 93.42          & 97.22                    \\ 
             QFace~\cite{sun2024task}       & $\times$        & $\times$          & $\times$             & $\times$          & $\times$          & $\times$                    \\ 
            Faceptor~\cite{qin2025faceptor}       & 99.40        & 96.34          & 93.65             & 94.75          & 92.27          & 95.28                    \\ 
            \textbf{FaceXFormer}    & 99.68        & 96.75          & 96.35             & 95.50          & 91.41          & 95.94                    \\ \hline
        \end{tabular}%
    }
    \vspace{-2pt}
    \caption{Performance comparison for face recognition.}
    \vspace{-12pt}
    \label{tab:fr}
\end{table}
Recent models such as SwinFace~\cite{qin2023swinface}, QFace~\cite{sun2024task}, and Faceptor~\cite{qin2025faceptor} also aim to unify multiple tasks but only address a subset of them. These approaches often exclude complex tasks such as segmentation, head pose estimation, and landmark prediction. Moreover, they rely on multiple decoders and computationally expensive attention mechanisms, adding to the overall computational overhead. In contrast, \textit{FaceXFormer} seamlessly unifies these complex tasks using a lightweight decoder and achieves state-of-the-art performance across them at a real-time FPS of $33.21$. It outperforms previous multi-task models in segmentation, head pose estimation, landmark prediction, attribute prediction, and face visibility prediction, while achieving the second-best performance in age estimation. In this work, we simultaneously train for ten heterogeneous tasks, posing a more formidable challenge than previous approaches. Despite this, \textit{FaceXFormer} effectively handles multiple tasks, achieving SOTA or competitive performance in real time. This success can be attributed to the efficiency of the proposed lightweight decoder, which employs a novel bi-directional cross-attention mechanism.

\subsection{Qualitative ``in-the-wild'' results}
\label{subsec:qualitative_results}
\begin{figure}[t]
    \centering
    \includegraphics[width=\linewidth]{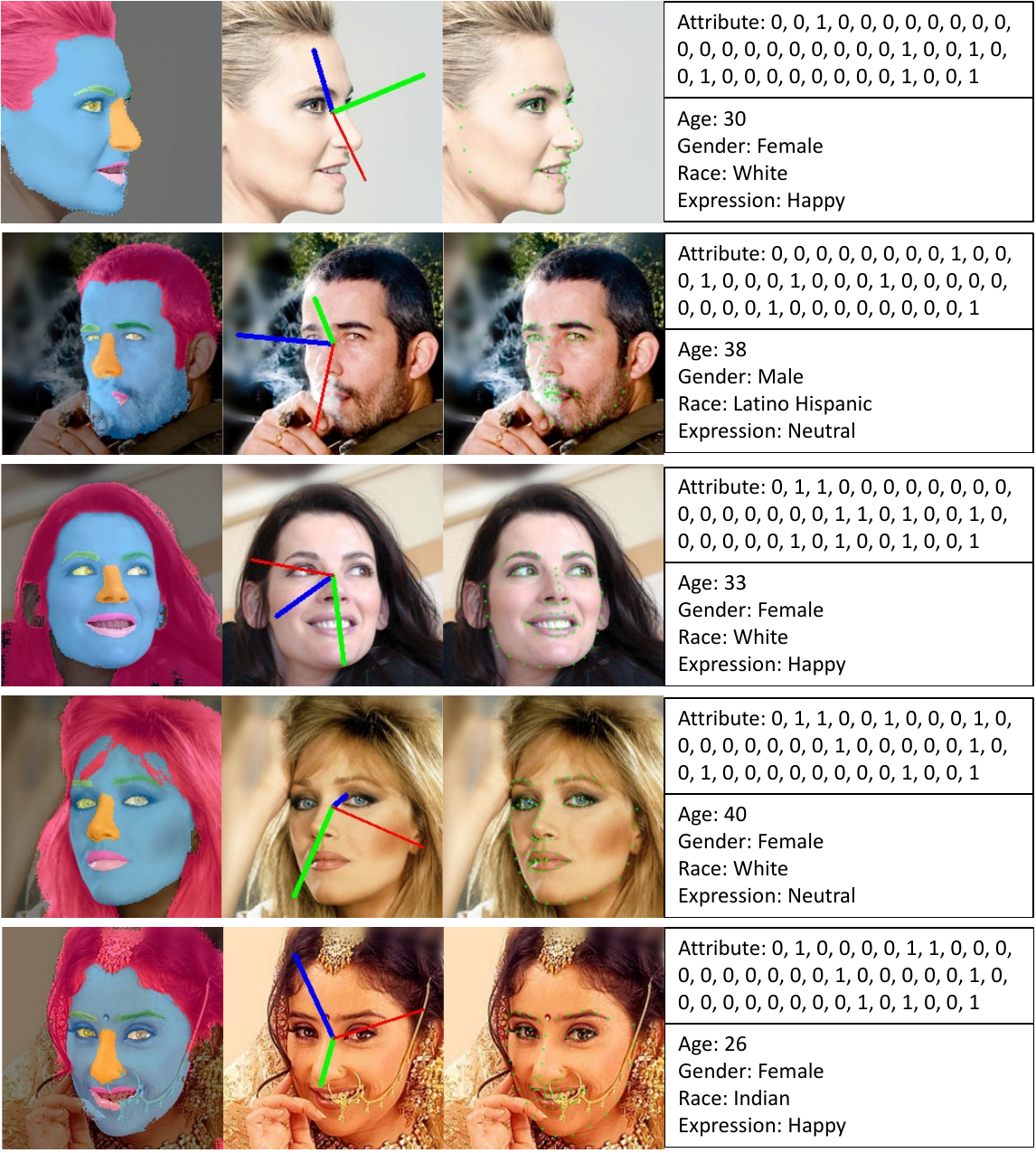}
    \vspace{-13pt}
    \caption{\textit{FaceXFormer} predictions on ``in-the-wild'' images}
    \vspace{-5pt}
    \label{fig:qualitative}
\end{figure}

In this section, we present the qualitative results of \textit{FaceXFormer} on randomly selected ``in-the-wild'' images. We select four random images and showcase the results for face parsing, head pose estimation, landmarks prediction, age estimation, gender and race classification, and attributes prediction in Figure~\ref{fig:qualitative}. Notably, the model successfully performs complex tasks such as face segmentation, head pose estimation, and landmark prediction, even when input samples exhibit extreme poses, occlusions, or blurring. Furthermore, \textit{FaceXFormer} can be effectively used as a tool to generate multiple annotations for each image, making it valuable for various downstream tasks. These results highlight \textit{FaceXFormer's} robust performance in challenging, real-world scenarios.

\section{Ablation Study}
\label{subsec:ablation}
\begin{table}[!ht]
\centering
\resizebox{\linewidth}{!}{%
\begin{tabular}{cccc|cccc}
\toprule
\multicolumn{1}{c}{MLP} & \multicolumn{1}{c}{Self} & \multicolumn{1}{c}{Cross} & \multicolumn{1}{c|}{Bi-dir.} 
& \multicolumn{1}{c}{HPE} & \multicolumn{1}{c}{Lnd} & \multicolumn{1}{c}{Attr.} & \multicolumn{1}{c}{Age} \\

\multicolumn{1}{c}{Fusion} & \multicolumn{1}{c}{Attn.} & \multicolumn{1}{c}{Attn.} & \multicolumn{1}{c|}{CA}
& \multicolumn{1}{c}{(MAE)} & \multicolumn{1}{c}{(NME)} & \multicolumn{1}{c}{(Acc.)} & \multicolumn{1}{c}{(MAE)} \\
\cmidrule(lr){1-4}\cmidrule(lr){5-8}
 & \checkmark & & \checkmark & 3.70 & 4.70 & 91.21 & 4.21 \\
\checkmark & \checkmark &  &  & 17.16 & 31.49 & 79.90 & 16.15 \\
\checkmark & \checkmark & \checkmark &  & 4.63 & 5.30 & 89.98 & 4.84 \\
\rowcolor[gray]{0.93} \checkmark & \checkmark &  & \checkmark & \textbf{3.52} & \textbf{4.67} & \textbf{91.83} & \textbf{4.17} \\
\bottomrule
\end{tabular}%
}
\vspace{-5pt}
\caption{Impact of various components on performance.}
\label{table:ablation}
\vspace{-10pt}
\end{table}

In this section, we explore the impact of different components of \textit{FaceXFormer} on performance. Additionally, we demonstrate that the proposed model exhibits minimal bias compared to other models by evaluating age and gender prediction across various demographics. Furthermore, we analyze the computational performance of different components of \textit{FaceXFormer} and compare it with existing multi-task models. Additional ablation studies on the impact of using different backbones of varying sizes are provided in the Appendix~\ref{sec:supp_ablation}.

\subsection{Impact of various components in FaceXFormer}
To evaluate the contribution of each component in \textit{FaceXFormer}, we conduct an ablation study focusing on the importance of specific design choices and their impact on performance across various tasks. The results of these experiments are summarized in Table~\ref{table:ablation}. We observe that without MLP fusion (row 1), there is a drop in performance, highlighting the importance of  of integrating multi-scale features to capture both global and local information essential for accurate predictions. The model performs extremely poorly (row 2) without cross-attention in the decoder, which is expected, as there is no interaction between face tokens and task tokens in this case. Introducing the proposed bi-directional cross-attention (row 4), which corresponds to \textit{FaceXFormer}, in the decoder provides a significant boost compared to using standard cross-attention (row 3), yielding improvements of $1.11$ MAE in head pose estimation, $0.63$ NME in landmark detection, $1.85$ accuracy points in attribute prediction, and $0.67$ MAE in age estimation. These results demonstrate the importance of MLP fusion and bi-directional cross-attention in the \textit{FaceXFormer} architecture.

\subsection{Bias Analysis and Ethical Considerations}
In our work, we utilize $17$ unique datasets for training and evaluation. We obtained these datasets following the procedures stated on their respective pages and signed the license agreements if and when necessary. As we train our models on multiple datasets designed for different tasks, the subjects across different age groups, genders, and races is not equal. This imbalance may introduce bias in the model. Therefore, we provide an analysis using the FairFace~\cite{karkkainen2021fairface} dataset, which is balanced in terms of age, gender and race. We follow~\cite{radford2021learning} and define the "Non-white" group to include multiple racial categories: "Black", "Indian", "East Asian", "Southeast Asian", "Middle Eastern" and "Latino". As can be seen from Table~\ref{tab:ethics}, \textit{FaceXFormer} shows the smallest performance discrepancy across different racial groups and exhibits minimal bias compared to other models despite being trained on fewer data points. This can be attributed to race estimation being the task in co-training.
\begin{table}[t]
\centering
\resizebox{\linewidth}{!}{
\begin{tabular}{@{}l@{\hspace{1em}}l@{\hspace{1em}}c@{\hspace{1em}}c@{\hspace{1em}}c@{\hspace{1em}}c@{\hspace{1em}}c@{}}
\toprule
 & Model & Data Points & White & Non-white & Average & Discrepancy \\ \hline
 \multirow{4}{*}{\rotatebox[origin=c]{90}{Age}} & FairFace & 100K & 60.05 & 60.63 & 60.52 & 0.58 \\ 
 & CLIP & 400M & 62.25 & 61.95 & 62.00 & -0.30 \\ 
 & FaRL & 20M & 61.49 & 61.84 & 61.78 & 0.35 \\  
 & FaceXFormer & 5.9M & 58.94 & 59.44 & 59.34 & 0.50 \\ \hline
\multirow{4}{*}{\rotatebox[origin=c]{90}{Gender}} & FairFace & 100K & 94.15 & 94.41 & 94.36 & 0.26 \\ \
 & CLIP & 400M & 94.87 & 95.78 & 95.61 & 0.91 \\ 
 & FaRL & 20M & 95.16 & 95.77 & 95.65 & 0.61 \\ 
 & FaceXFormer & 5.9M & 95.34 & 95.19 & 95.22 & -0.09 \\ \bottomrule
\end{tabular}}
\vspace{-5pt}
\caption{Age and gender accuracy w.r.t race groups on FairFace}
\label{tab:ethics}
\vspace{-8pt}
\end{table}

\subsection{Computational Performance Analysis}
We present a computational performance analysis of the proposed method compared to previous multi-task models in Table~\ref{tab:fps} to highlight its efficiency. \textit{FaceXFormer} achieves the fastest inference speed among multi-task face analysis models, with an FPS of $33.2$ (FP32) and $100.1$ (FP16), outperforming previous multi-task model Faceptor~\cite{qin2025faceptor}. This improvement is attributed to the proposed \textit{FaceX} decoder, which employs a novel bi-directional cross-attention mechanism, enabling \textit{FaceXFormer} to maintain only two decoder layers while ensuring effective face feature extraction. Moreover, \textit{FaceXFormer} significantly reduces computational cost, requiring only $114$ GFLOPs compared to $167$ GFLOPs in Faceptor, leading to a substantial reduction in latency from $69.9$ ms to $30.1$ ms in FP32 and from $23.7$ ms to $10.0$ ms in FP16. With its reduced computational cost and faster inference, \textit{FaceXFormer} achieves state-of-the-art performance across most tasks, demonstrating the effectiveness of its lightweight yet powerful design.
\begin{table}[h]
    \centering
    \vspace{-2pt}
    \resizebox{0.47\textwidth}{!}{
    \begin{tabular}{@{}llccccc}
        \toprule
        \multirow{2}{*}{\textbf{Component}} 
        & \multirow{2}{*}{\textbf{Model}} 
        & \multirow{2}{*}{\textbf{FLOPs}} 
        & \multicolumn{2}{c}{\textbf{Latency (ms)}} 
        & \multicolumn{2}{c}{\textbf{FPS}} \\
        \cmidrule(lr){4-5} \cmidrule(lr){6-7}
        & & \textbf{(GFLOPs)} 
        & \textbf{FP32} & \textbf{FP16} 
        & \textbf{FP32} & \textbf{FP16} \\
        \midrule
        \multirow{2}{*}{Backbone} 
         & Faceptor~\cite{qin2025faceptor}     & 64 & 19.1 & 6.5 & 52.3  & 153.8 \\
         & \textit{\textbf{FaceXFormer}}  & 47   & 12.5 & 4.0 & 80.0  & 250.0 \\
        \midrule
        \multirow{2}{*}{Decoder} 
         & Faceptor~\cite{qin2025faceptor}      & 85   & 34.0 & 11.5 & 29.4  & 87.0  \\
         & \textit{\textbf{FaceXFormer}}  & 55   & 13.6 & 4.5 & 73.5  & 222.2 \\
        \midrule
        \multirow{2}{*}{Task Heads} 
         & Faceptor~\cite{qin2025faceptor}      & 18   & 6.5 & 2.2 & 153.8 & 454.5 \\
         & \textit{\textbf{FaceXFormer}}  & 12   & 4.0 & 1.5 & 250.0 & 666.7 \\
        \midrule
        \multirow{2}{*}{\begin{tabular}[l]{@{}l@{}}Total Inference\\Time\end{tabular}}
         & Faceptor~\cite{qin2025faceptor}      & 167 & 69.9 & 23.7 & 14.3  & 42.2  \\
         & {\cellcolor[gray]{0.93}\textit{\textbf{FaceXFormer}}}
           & {\cellcolor[gray]{0.93}114}
           & {\cellcolor[gray]{0.93}30.1}
           & {\cellcolor[gray]{0.93}10.0}
           & {\cellcolor[gray]{0.93}\textcolor{red}{\textbf{33.2}}}
           & {\cellcolor[gray]{0.93}\textcolor{red}{\textbf{100.1}}} \\
        \bottomrule
    \end{tabular}
    }
    \vspace{-3pt}
    \caption{Computational performance: \textit{FaceXFormer} vs Faceptor.}
    \label{tab:fps}
    \vspace{-13pt}
\end{table}

\section{Conclusion}
\label{sec:conclusion}
\textit{FaceXFormer} introduces a novel end-to-end unified model that efficiently handles a wide range of facial analysis tasks in real-time. By adopting a transformer-based encoder-decoder architecture and representing each task as a learnable token, our approach seamlessly integrates multiple tasks within a single framework while maintaining minimal computational cost and fast inference times. The proposed lightweight decoder, \textit{FaceX, incorporates a novel bi-directional cross-attention mechanism, enhancing the model's ability to learn robust and} generalized face representations across diverse tasks. Comprehensive experiments demonstrate that \textit{FaceXFormer} achieves state-of-the-art performance across multiple facial analysis tasks, achieving a real-time FPS of $33.21$. In broader applications, \textit{FaceXFormer} can serve as an annotator for large-scale face datasets and can be integrated into existing facial analysis systems to provide extra information, making it a valuable tool for surveillance, subject analysis, and image retrieval.

\section*{Acknowledgment}
This research is based upon work supported in part by the Office of the Director of National Intelligence (ODNI), Intelligence Advanced Research Projects Activity (IARPA), via [2022-21102100005]. The views and conclusions contained herein are those of the authors and should not be interpreted as necessarily representing the official policies, either expressed or implied, of ODNI, IARPA, or the U.S. Government. The US Government is authorized to reproduce and distribute reprints for governmental purposes notwithstanding any copyright annotation therein.
{
    \small
    \bibliographystyle{ieeenat_fullname}
    \bibliography{main}
}

\appendix
\onecolumn 
\makebox[\textwidth][c]{\Large \textbf{Appendix}}

\counterwithin{figure}{section}
\counterwithin{table}{section}

\section{Overview}
\label{sec:supp_intro}

As part of the Appendix, we present the following as an extension to the ones shown in the paper:

\begin{itemize}
    \item Broader Impact (Section~\ref{sec:discussion})
    \item Ablation study  (Section~\ref{sec:supp_ablation})
    \item Cross-dataset Evaluation (Section~\ref{sec:supp_cross})
    \item In-the-wild Visualization (Section~\ref{sec:supp_vis})
    \item Dataset details (Section~\ref{sec:supp_implementation})
\end{itemize}

\section{Discussion}
\label{sec:discussion}
The world is moving towards transformers because of its potential to model large amounts of data~\cite{brown2020language, kirillov2023segment, videoworldsimulators2024}. Presently, the face community lacks large-scale annotated datasets to train foundational models capable of performing a wide spectrum of facial tasks. The largest clean dataset, WebFace42M~\cite{zhu2021webface260m}, lacks annotations for face parsing, landmarks detection, headpose, expression, race and facial attributes. \textit{FaceXFormer} can be used as an annotator for large-scale data, and can be continually improved through successive rounds of annotation and fine-tuning. We aim to propel the face community towards developing foundation models that cater to a variety of facial tasks. Additionally, \textit{FaceXFormer} is a lightweight model that provides real-time output based on task-specific queries and can be appended with existing facial systems to provide additional information. It can also serve as a valuable tool in surveillance, and provide auxiliary information for subject analysis and image retrieval.

\section{Ablation Study} 
\label{sec:supp_ablation}

To evaluate the impact of different backbones on performance and FPS, we conduct an ablation study comparing various backbone architectures in \textit{FaceXFormer}. We categorize head pose estimation, landmark prediction, and age estimation as regression (Reg) tasks, while attribute prediction and facial expression recognition fall under classification (Cls). Additionally, face parsing is denoted as segmentation (Seg). The results of these experiments are summarized in Table~\ref{tab:backbone}.
\begin{table}[!h]
    \centering
    \resizebox{0.4\columnwidth}{!}{\begin{tabular}{l@{\hspace{1em}}ccccc@{}}
        \toprule
        \textbf{Backbone} & \textbf{Seg} & \textbf{Reg} & \textbf{Cls} & \textbf{FPS} & \textbf{Params} \\ \midrule
            MobileNet & 91.21 & 4.64 & 88.22 & 39.76 & 25.32 \\
            ResNet101  & 91.49 & 4.37 & 88.91 & 34.98 & 65.54 \\ 
            ConvNext-B & 92.08 & 4.35 & 89.09 & 36.61 & 110.19 \\  
            Swin-B & 92.01 & 4.12 & 90.03 & 33.21 & 109.29 \\ 
            \bottomrule
    \end{tabular}}
    \caption{Effect of different backbones on performance and FPS.}
    \label{tab:backbone}

\end{table}

\noindent From the results, we observe that ConvNeXt achieves the best performance in segmentation with an F1 score of $92.08$\%. The Swin Transformer backbone excels in both regression and classification tasks, with a mean error of $4.12$ and a mean accuracy of $90.03$\%, respectively. In contrast, MobileNet demonstrates the lowest performance metrics, including an F1 score of $91.21$\% and a mean error of $4.64$, highlighting its limitations in handling larger, more complex datasets due to its smaller receptive field compared to the Swin Transformer. The selection of the Swin Transformer as the backbone for \textit{FaceXFormer} is driven by its superior scalability and global contextual understanding, both of which are essential for facial analysis tasks.

\section{Cross-Dataset Evaluation}
\label{sec:supp_cross}
We conduct additional cross-dataset experiments to demonstrate the effectiveness of \textit{FaceXFormer} in scenarios that closely resemble real-life conditions. These scenarios involve previously unseen, unconstrained face images characterized by significant variability in background, lighting, pose, and other factors. As shown in Table~\ref{table:cross_dataset}, \textit{FaceXFormer} outperforms the existing state-of-the-art model, STARLoss~\cite{zhou2023star}, on the 300VW dataset. This highlights \textit{FaceXFormer}'s effectiveness in landmark detection under in-the-wild scenarios. The cross-dataset results support the rationale presented in this paper: the necessity of a unified facial analysis model capable of performing multiple tasks on unconstrained, in-the-wild faces, particularly for real-time applications. \textit{FaceXFormer} addresses this gap and achieves state-of-the-art performance.

\begin{table}[ht]
\centering
\resizebox{0.75\textwidth}{!}{
    \begin{tabular}{@{}l@{\hspace{1em}}c@{\hspace{1em}}c@{\hspace{1em}}c@{\hspace{1em}}c@{}}
        \toprule 
        \multirow{2}{*}{Method} & 300VW (Cat.A)& 300VW (Cat.B) & 300VW (Cat.C) & LFWA (Gender) \\ \cmidrule{2-5}
        & NME & NME & NME & Acc. \\ \midrule
        PANDA~\cite{zhang2014panda} & - & - & - & 92.00\\
        STARLoss~\cite{zhou2023star} & 3.97 & \textbf{3.39} & 8.42 & -\\ \midrule
        \textit{\textbf{FaceXFormer}} & \textbf{3.90} & 3.58 & \textbf{6.75} & \textbf{92.74}\\ \bottomrule
    \end{tabular}}
\caption{Cross Dataset evaluation of \textit{FaceXFormer}.}
\label{table:cross_dataset}
\end{table}

\section{In-the-wild Visualization}
\label{sec:supp_vis}
\begin{figure*}[t]
  \centering
    \includegraphics[width=1.0\linewidth]{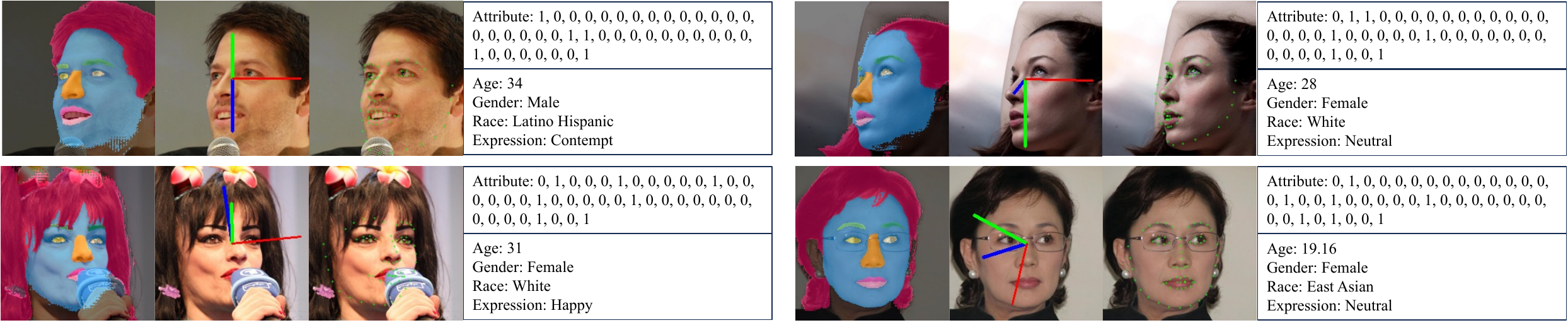}
   \caption{Visualization of ``in-the-wild" images for multiple tasks. Attributes represent the $40$ binary attributes defined in the CelebA~\cite{liu2015faceattributes} dataset, indicating the presence ($1$) or absence ($0$) of specific facial attributes.}
   \label{fig:supp_viz}
\end{figure*}

We randomly selected images from the web and treated them as "in-the-wild" images. The qualitative results for all tasks are presented in Figure~\ref{fig:supp_viz}. Our observations indicate that \textit{FaceXFormer} produces promising results even in the presence of occlusions, extreme angles, and accessories.

\section{Datasets and Implementation Details}
\label{sec:supp_implementation}
In this section, we detail the dataset characteristics and the augmentations applied to each dataset during training. \textit{FaceXFormer} is trained using multiple datasets, which have varying sample sizes. Datasets with a larger number of images may dominate the training process and create bias. To mitigate this, we employ upsampling to ensure that each batch is represented by samples from every dataset. This is achieved by repeating the samples of smaller datasets through upsampling and then randomly sampling images from the upsampled set. The model is trained for $12$ epochs with a total batch size of $384$ and an initial learning rate of $1e^{-4}$, which decays by a factor of $10$ at the $6^{th}$ and $10^{th}$ epochs. We use the AdamW optimizer with a weight decay of $1e^{-5}$ for gradient updates. 

\subsection{Face Parsing}
We use CelebAMask-HQ~\cite{CelebAMask-HQ} for training and evaluation of \textit{FaceXFormer}. CelebAMask-HQ contains 30,000 high-resolution face images annotated with 19 classes. The classes used for training and evaluation include: skin, face, nose, left eye, right eye, left eyebrow, right eyebrow, upper lip, mouth, and lower lip. During training, we resize the images to $224 \times 224$, before feeding them into the model.

\subsection{Landmarks Detection}
We utilize the 300W dataset~\cite{sagonas2013300} for the training and evaluation of \textit{FaceXFormer}. The 300W dataset contains 3,148 images in its training set and 689 test images, which are categorized into three overlapping test sets: common (554 images), challenge (135 images), and full (689 images). It encompasses a wide variety of identities, expressions, illumination conditions, poses, occlusions, and face sizes. All images are annotated with 68 landmark points. For cross-dataset testing of multi-task methods, we employ the 300VW dataset~\cite{shen2015first}. This dataset provides three test categories: Category-A (well-lit conditions, comprising 31 videos with 62,135 frames), Category-B (mildly unconstrained conditions, consisting of 19 videos with 32,805 frames), and Category-C (challenging conditions, including 14 videos with 26,338 frames). We report the results for all three categories. During training, we apply various data augmentations such as random rotation ($\pm 18^\circ$), random scaling ($\pm 10\%$), random translation ($5\% \times 224$), random horizontal flip ($50\%$), random gray ($20\%$), random Gaussian blur ($30\%$), random occlusion ($40\%$) and random gamma adjustment($20\%$). Additionally, we align the images using five landmarks points. 

\subsection{Head Pose Estimation}
We utilize the 300W-LP dataset~\cite{zhu2016face}, which contains approximately 122,000 samples. For performance evaluation, we use the BIWI dataset~\cite{fanelli2013random}, comprising 15,678 images of 20 individuals (6 females and 14 males, with 4 individuals recorded twice). The head pose range spans approximately $\pm 75^\circ$ yaw and $\pm 60^\circ$ pitch. During training, we loosely crop the face images based on the landmarks and apply several augmentations, including random gray ($10\%$), random Gaussian blur ($10\%$), random resized crop ($80\% to 100\%$)and random gamma adjustment($10\%$).

\subsection{Attributes Prediction}
We utilize the CelebA~\cite{liu2015faceattributes} dataset for training and the LFWA~\cite{wolf2010effective} dataset for cross-dataset evaluation of multi-task methods. CelebA comprises 202,599 facial images, each annotated with 40 binary labels that indicate various facial attributes such as hair color, attractive, bangs, big lips, and more. LFWA consists of 13,143 facial images, annotated with the same set of facial attributes. During training, we apply several augmentations, including random rotation ($\pm 18^\circ$), random scaling ($\pm 10\%$), random translation ($1\% \times 224$), random horizontal flip ($50\%$), random gray ($10\%$), random Gaussian blur ($10\%$), and random gamma adjustment($20\%$). 

\subsection{Age/Gender/Race Estimation}
We utilize the FairFace~\cite{karkkainenfairface} and UTKFace~\cite{zhang2017age} datasets for training, and the FFHQ~\cite{karras2019style} dataset for cross-dataset testing. FairFace comprises 108,501 images, balanced across seven racial groups: White, Black, Indian, East Asian, Southeast Asian, Middle Eastern, and Latino. The UTKFace dataset contains 20,000 facial images annotated with age, gender, and race. In our work, we follow the 'race-4' annotation scheme, categorizing individuals into five racial labels: White, Black, Indian, Asian, and Others. Age annotations are categorized into decade bins: 0--9, 10--19, 20--29, 30--39, 40--49, 50--59, 60--69, and over 70. Gender is annotated with two labels: male and female. Additionally, we incorporate the MORPH-II dataset~\cite{ricanek2006morph}, which contains 55,134 facial images of 13,617 subjects aged between 16 and 77 years. This dataset provides annotations for age, gender, and race, with a predominance of male subjects and a significant representation of Black and White individuals. For age estimation tasks, we train on both UTKFace and MORPH-II datasets and evaluate our models on the MORPH-II dataset to assess performance. During training, we apply augmentations such as random rotation ($\pm 18^\circ$), random scaling ($\pm 10\%$), random translation ($1\% \times 224$), random horizontal flip ($50\%$), random grayscale conversion ($10\%$), random Gaussian blur ($10\%$), and random gamma adjustment ($10\%$).

\subsection{Facial Expression Recognition}
We utilize the RAF-DB~\cite{li2017reliable} and AffectNet~\cite{mollahosseini2017affectnet} datasets for training and RAF-DB~\cite{li2017reliable} dataset for intra-dataset evaluation. RAF-DB is a facial expression dataset with approximately 30,000 images. The dataset includes variability in subjects' age, gender, ethnicity, head poses, lighting conditions, and occlusions (e.g., glasses, facial hair, or self-occlusion). RAF-DB provides annotations for seven basic emotions that are surprise, fear, disgust, happiness, sadness, anger, and neutral. AffectNet is one of the largest facial expression datasets with approximately 440,000 images that are manually annotated for the presence of eight discrete facial expressions: neutral, happy, angry, sad, fear, surprise, disgust, contempt. During training, we apply augmentations such as random rotation ($\pm 18^\circ$), random scaling ($\pm 10\%$), random translation ($1\% \times 224$), random horizontal flip ($50\%$), random grayscale conversion ($10\%$), random Gaussian blur ($10\%$), random color jitter ($10\%$), and random gamma adjustment ($10\%$).

\subsection{Face Recognition}
We utilize the MS1MV3~\cite{guo2016ms} dataset for training our face recognition models and evaluate their performance using LFW~\cite{huang2008labeled}, CFP-FP~\cite{cfp-paper}, AgeDB~\cite{moschoglou2017agedb}, CALFW~\cite{zheng2017cross}, and CPLFW~\cite{zheng2018cross}. MS1M-V3 is a cleaned version of the MS-Celeb-1M dataset, containing approximately 5.1 million images of 93,000 identities, making it suitable for large-scale face recognition training. For evaluation, LFW (Labeled Faces in the Wild) consists of 13,233 images of 5,749 individuals and is designed for face verification in unconstrained environments. CFP-FP (Celebrities in Frontal-Profile) contains 7,000 images of 500 subjects and focuses on frontal-to-profile face verification. AgeDB provides 12,240 images of 440 subjects, spanning ages from 3 to 101 years, to evaluate age-invariant face verification. CALFW (Cross-Age LFW) introduces age variations by selecting positive pairs with large age gaps and negative pairs with similar age, race, and gender attributes. CPLFW (Cross-Pose LFW) is derived from LFW and emphasizes pose variation by selecting positive pairs with different poses and negative pairs with similar pose, race, and gender. These datasets collectively cover diverse challenges, including pose, age, and other variations, enabling a comprehensive evaluation of face recognition models. We do not apply any augmentations during training but preprocess images by aligning them based on five facial keypoints before feeding them into the model.

\subsection{Visibility Prediction}
We utilize the COFW~\cite{burgos2013robust} dataset, which is annotated with 29 landmarks for landmarks visibility prediction. Each landmark is associated with 29 binary labels that indicate its visibility. We loosely crop the faces and apply augmentations, including random horizontal flip ($50\%$), random gray ($10\%$), random Gaussian blur ($10\%$), and random gamma adjustment($10\%$).

\end{document}